\documentclass[journal]{IEEEtran}

\ifCLASSINFOpdf
   \usepackage[pdftex]{graphicx}
   \graphicspath{{../fig/}}
   \DeclareGraphicsExtensions{.pdf,.jpeg,.png,.jpg,.tikz,.gif}
\else
\fi
\usepackage{amssymb}
\usepackage[cmex10]{amsmath}
\usepackage{algorithmic}
\usepackage{algorithm}
\usepackage[tight,footnotesize]{subfigure}
\usepackage{color}
\usepackage{dblfloatfix}
\usepackage[squaren,Gray]{SIunits}

\begin{document}
%%%%%%%%%%%%%%%%%%%%%%%%%%%%%%%%%%%%%%%%%%%%%%%%%%%%%%%%%%%%%%
\title{ Event-Based Structured Light for Depth Reconstruction using Frequency Tagged Light Patterns\\
}
\author{
T. Leroux, S.-H. Ieng and R. Benosman 

%Natural computation group
University of Pittburgh, Carnegie Mellon Univeristy, Sorbonne Universitas\\
benosman@pitt.edu
%\and Xavier Lagorce
%\and Sio-Hoi Ieng
%\and Ryad Benosman
}
%%%%%%%%%%%%%%%%%%%%%%%%%%%%%%%%%%%%%%%%%%%%%%%%%%%%%%%%%%%%%%
\maketitle

%%%%%%%%%%%%%%%%%%%%%%%%%%%%%%%%%%%%%%%%%%%%%%%%%%%%%%%%%%%%%%
\begin{abstract}
This paper presents a new method for 3D depth estimation using the output of an asynchronous time driven image sensor. In association with a high speed Digital Light Processing projection system, our method achieves real-time reconstruction of 3D points cloud, up to several hundreds of hertz. Unlike state of the art methodology, we introduce a method that relies on the use of frequency tagged light pattern that make use of the high temporal resolution of event based sensors. This approch eases matching as each pattern unique frequency allow for any easy matching between displayed patterns and the event based sensor. Results are show on real scenes.
\end{abstract}
%%%%%%%%%%%%%%%%%%%%%%%%%%%%%%%%%%%%%%%%%%%%%%%%%%%%%%%%%%%%%%


% \fi
\IEEEpeerreviewmaketitle

%%%%%%%%%%%%%%%%%%%%%%%%%%%%%%%%%%%%%%%%%%%%%%%%%%%%%%%%%%%%%%
\section{Introduction}
%%%%%%%%%%%%%%%%%%%%%%%%%%%%%%%%%%%%%%%%%%%%%%%%%%%%%%%%%%%%%%
\subsection{Motivation}
\IEEEPARstart{S}{tructured light} is considered the most reliable technique to estimate depth and is behind most commonly used methods in the field of non-contact 3D measurement. Because of its accuracy and simplicity, its the method of choice in many applications including autonomous vehicle navigation\cite{Escalera1996, Asada1990}, factory inspection processes \cite{Kakinoki1990, Chin1988}, object recognition\cite{Petriu2000, Park2001} or reverse engineering \cite{Page2003}. The term \emph{structured light} itself refers to the projection of simple or encoded light patterns onto the illuminated scene \cite{Klette1998}. Its main purpose is to simplify the extraction of features in the image of the scene captured by a camera, and facilitate the pairing process. Like in classic stereoscopic systems, the measure of 3D information using structured light is done by triangulation when corresponding points are identified. 

This paper presents a method based on a completely new neuromorphic vision sensor, whose dynamic and temporal properties are unmatched. Using a totally different approach of vision, it allows to push back the limitations of former systems.
Uses of structured light technology can be tracked back to the early 80's \cite{Batlle1998}, but its only since the developement of the Kinect sensor by Microsoft in 2008 that structured light became affordable.
%, mostly due to the aggressive market strategy of the brand
A large number of devices have emerged since. However, most of them share two major limitations resulting from the use of classic frame-grabbing cameras.
Because they have short dynamic ranges, captured intensity of the projected signal can be extremely low in strong ambient illumination, thus resulting in poor depth estimations. This is especially true in outdoor environnement where sunlight is often 2-5 orders of magnitude brighter than the projected light. Another contrainst is the frequency at which classic cameras operate (between 30 and $120\hertz$ for commonly used cameras), which is not suited to follow and reconstruct fast moving objects. Some of the latest methods using high speed frame-grabbing cameras achieve 3D reconstruction up to $500\hertz$ but at high energy and data cost. 

Recent developement in neuromorphic vision opens up new alternatives to classic frame-grabbing cameras and a new way to process visual information. 
The presented system uses an biomimetic artificial retina with a 143dB dynamic range able to acquire asynchronous data at a microsecond precision. 

\subsection{State of the Art}
Structured light techniques differs from one another essentially because of the encoding method used to create the projected pattern, or sequence of patterns. In the litterature, several comparative studies can be found that analyse pattern encoding strategies over the last decades \cite{Su2001, Salvi2004, Salvi2010}. Methods based on multiple patterns (i.e time multiplexing) are generally better when dealing with static objects and offers highly accurate measurements. Posdamer et al. \cite{Posdamer1982} used a succession of patterns generating a pure binary code, followed by works to decrease errors at brightness boundaries \cite{Inokuchi1984}, increase the resolution \cite{Bergmann1995} or reduce the number of frame needed \cite{Caspi1998}. For these methods, a high resolution is obtained as depth information is calculated at every pixel. Noise is mostly generated by the movement of objects between successive patterns of the projection. Other methods based on single pattern projection, using spatial multiplexing, are more robust for a moving object. Those includes works on patterns composed of multiple slits of different sizes \cite{Maruyama1993, Durdle1998} or colors \cite{Zhang2002, Salvi1998}, patterns using specific shapes or colors organized in pseudorandom arrays \cite{Albitar2007, Maurice2011} or direct coding patterns in which each pixel is associated with a unique codeword by either spatial grading \cite{Tajima1990} or phase shifting \cite{Guan2003, Ono2004}. Recently, work has been done aiming to reach real-time reconstruction. Most of these attempts were made using a high speed frame based camera coupled with a fast DLP projector to allow projection of binary patterns at speeds over $1\kilo\hertz$  \cite{Ishii2007, Zhang2010}. These methods use both spatial and time multiplexing to get the best from both worlds and achieve up to $500\hertz$ reconstructions, while the current average reconstruction rate is somewhere between 30-100\hertz.

%%%%%%%%%%%%%%%%%%%%%%%%%%%%%%%%%%%%%%%%%%%%%%%%%%%%%%%%%%%%%%
\section{Hardware}
%%%%%%%%%%%%%%%%%%%%%%%%%%%%%%%%%%%%%%%%%%%%%%%%%%%%%%%%%%%%%%
\subsection{Time encoded imaging}
\label{sec:siliconRetina}
Biomimetic event-based cameras are a novel type of vision devices that - like their biological counterparts - are driven by "events" happening within the scene. They are not like conventional vision sensors
which, by contrast, are driven by artificially created timing and control signals (e.g. frame clock) which have no relation whatsoever to the source of the visual information \cite{Lichtsteiner2008}. Over the
past few years, a variety of these event-based devices have been implemented, including temporal contrast vision sensors that are sensitive to relative luminance change \cite{Posch2011, Leero2011}, gradient-based sensors sensitive to
static edges \cite{Andreou1991}, edge-orientation sensitive devices and optical-flow sensors \cite{Stocker2004}. Most of these vision sensors output visual information about the scene in the form of asynchronous address events (AER)
\cite{Boahen2000} and encode the visual information in the time dimension and not as voltage, charge or current. The presented depth estimation method is designed to work on the data
delivered by such a time-encoding sensor.
%and  to take full advantage of the high temporal resolution and the sparse data representation.

The ATIS used in this work is a time-domain encoding vision sensors with $304 \times 240$ pixels resolution. \cite{Lichtsteiner2008}. The sensor contains an array of fully autonomous pixels that combine an illuminance relative change detector
circuit and a conditional exposure measurement block.

\begin{figure}[!ht]
\centering
\includegraphics[width=1\columnwidth]{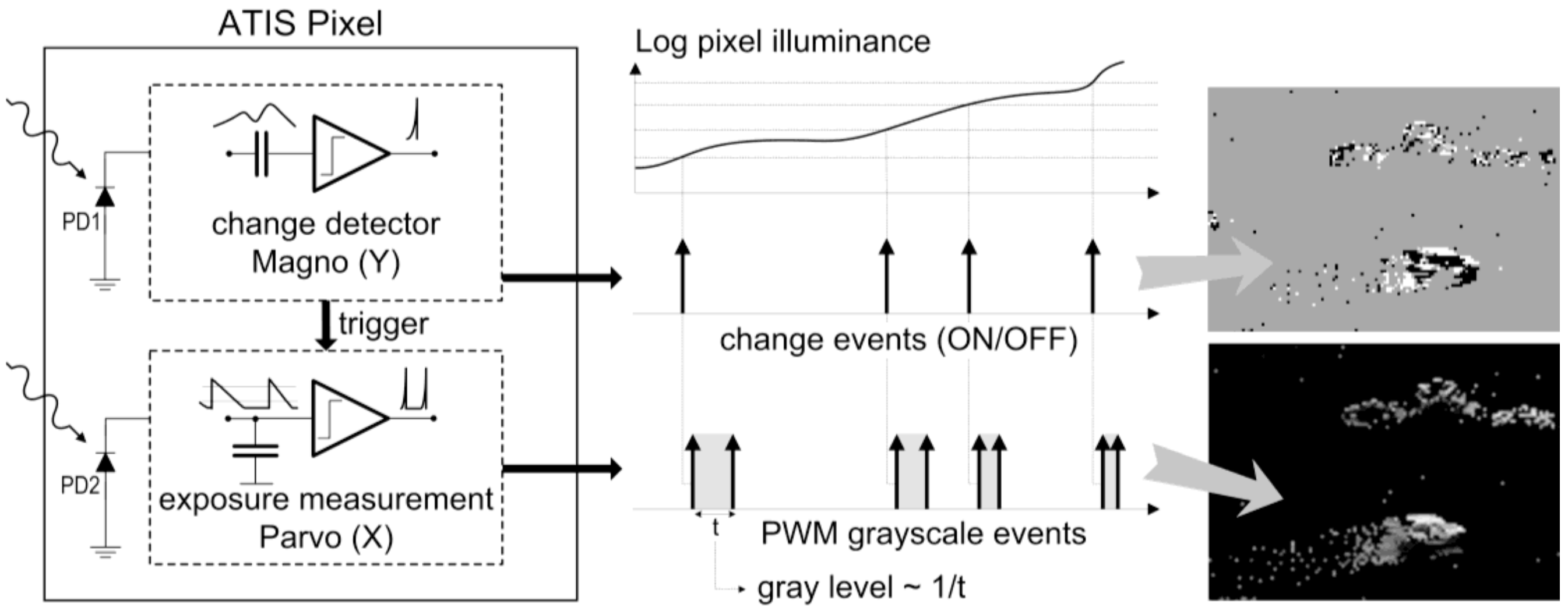}
\caption{Functional diagram of an ATIS pixel \cite{Posch2011}. Two types of asynchronous events, encoding change and brightness information, are generated and transmitted individually by each pixel in the imaging array.}
\label{fig:atis}
\end{figure}

As shown in the functional diagram of the ATIS pixel in Fig.~\ref{fig:atis}, the relative change detector individually and asynchronously initiates the measurement of an exposure/gray scale value only if
- and immediately after - a brightness change of a certain magnitude has been detected in the field-of-view of the respective pixel. The exposure measurement circuit in each pixel individually
encodes the absolute instantaneous pixel illuminance into the timing of asynchronous event pulses, represented as inter-event intervals.

Since the ATIS is not clocked like a conventional camera, the timing of events can be conveyed with a very accurate temporal resolution in the order of microseconds. The time-domain encoding of the intensity information automatically optimizes the exposure time separately for each pixel instead of imposing a fixed integration time for the entire array, resulting in an exceptionally high dynamic range and improved signal to noise ratio. The pixel-wise change detector driven operation yields almost ideal temporal redundancy suppression, resulting in a maximally sparse encoding of the image data.

\subsection{Digital Light Processing}

\begin{figure}[!ht]
\centering
\includegraphics[width=1\columnwidth]{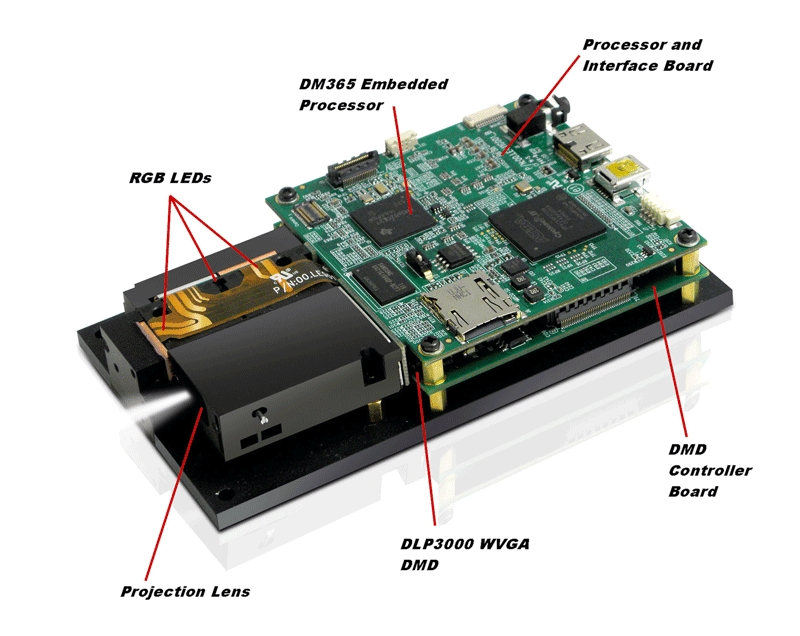}
\caption{The DLP lightcrafter module \textregistered}
\label{fig:DMD}
\end{figure}

Digital Light Processing (DLP) is an image projection technology originally developed by Texas Instrument, relying heavily on a chip containing orientable micromirrors called Digital Micromirror Device (DMD). A light bulb sends light on the chip's mirror array that reflect light back in a lens to generate an image on the screen. DMD was invented by Hornbeck and Nelson in 1987.
%% It is a chip composed of a rectangular pixel array of microscopic mirrors that can be individually rotated to either $+10\degree$ or $-10\degree$, corresponding to the pixels of a binary image. Each mirror is around $16\micro\meter$ across and is mounted on a yoke connected to torsion hinges. The mirror position is controled by electrostatic attraction with two pairs of electrode that maintain the mirror in its current state when bias charges are applied. To move the mirrors, desired states are loaded into an SRAM memory cell for each pixel. When every cells are loaded, bias voltage is removed and charges from the SRAM prevail, moving the mirror. Restoring the bias voltage hold the mirror in position and the next state is loaded in memory.
In our setup we use a DLP lightcrafter model (Fig.~\ref{fig:DMD}) which is a complete projection system embedding a WVGA micromirror array. It is able to project binary frames of $304 \times 240$ pixels at a $0.7\milli\second$ time resolution. Faster and bigger models exist, but this model benefits from the implementation of an asynchronous control driver that enables to manipulate up to $5000$ specific mirrors per time step instead of having to modify the complete pixel array synchronously.
%% This system allows a switching rate for binary frames up to $10\kilo\hertz$ for a $1080\times 1920$ pixels DMD array.

\subsection{Setup}
\begin{figure}[!ht]
\centering
\includegraphics[width=0.8\columnwidth]{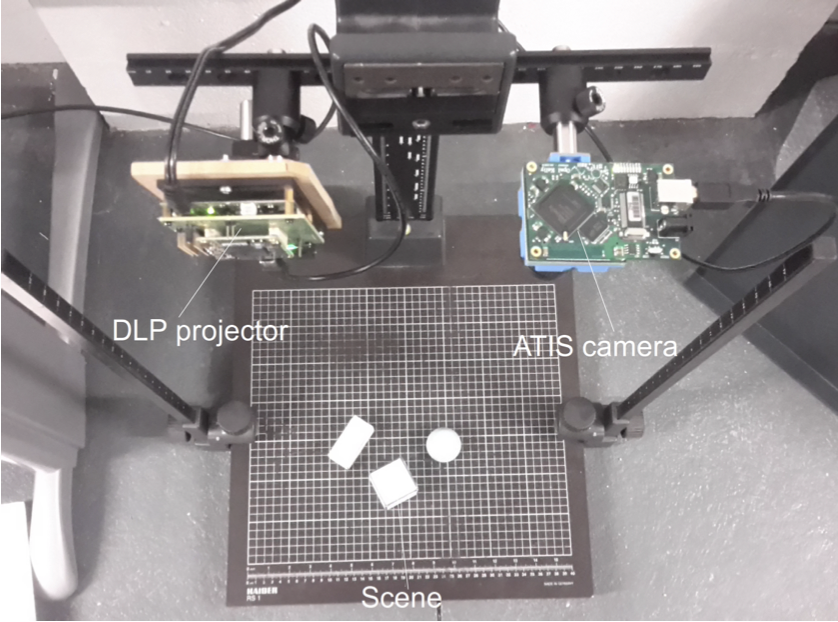}
\caption{Experimental setup}
\label{fig:setup}
\end{figure}

For the experiments in this paper we developed an active stereoscopic vision system using the association of an ATIS camera and a DLP projector (Fig.~\ref{fig:setup}). With its very high frame rate, the DMD used to produce binary projection can make profit of the high temporal resolution of the ATIS. An event is conveyed to the output buffer in a few hundred nanoseconds. This means that the time needed to change the position of DMD mirrors is fast enough to obtain spatially and temporally dense information. However limitation arises when exposing the whole pixel array of the ATIS to high frequency changes. An estimation of the maximal event flow the ATIS can currently manage is around 8 mega events per second. Even if this limitation will probably be overcome in the future, a $1\kilo\hertz$ stimulation of the sensor is limited to 800 events per step which represents only $1\%$ of the camera's pixel array. Due to this limitation a tradeoff has to be made between temporal and spatial resolution.

%%%%%%%%%%%%%%%%%%%%%%%%%%%%%%%%%%%%%%%%%%%%%%%%%%%%%%%%%%%%%%
\section{Pattern codification strategies}
%%%%%%%%%%%%%%%%%%%%%%%%%%%%%%%%%%%%%%%%%%%%%%%%%%%%%%%%%%%%%%

To get the better out of the ATIS sensor, it is mandatory to choose a suitable pattern codification. In the event domain, the only information available is the polarity and timestamp of pixels, thus the correspondance problem can't be solve by color or grey level coding. Moreover, the need for an asynchronous system impose that our pattern can be segmented in small independent patches that we could use to extract local information only when it is needed. This constraint has a direct impact on the spatial organisation of the pattern and direct methods like the one used in \ref{NorthwesternUni2015} must be discarded. In the next section we expand over three different methods that could be used to gather local data independently as well as being decodable without too much computation from the sensor's output to ensure a decent reconstruction speed. The first method we will develop is based on the well known time-multiplexing binary coding (refs?) but encode the binary frames as a signal with set frequency and dutycycle, each element is then part of a spatial codification that can be decoded locally.
Our second method aim to make use of spatio-temporal continuity of events generated from fast moving objects to create codewords based on spatial orientation of multiple moving dots. This approach differs only from the first in the way information is extracted from the sensor, but the decoding process is done the same way. Lastly, we developed an approach based on phase-shifting algorithms, but instead of evaluating the phase from several shifted patterns (ref), we continuously move a series of stripes and record precise time difference between stripe positions from an arbitrary origin, which gives us a robust measurement. This last method is able to gather information at the pixel level which is more accurate than previous ones but in exchange the pattern can only be segmented roughly for asynchronous projection.

%%The objective of the pattern codification is to find an efficient coding method to extract the 3D information under some fixed constraints. Those constraints could be of different type (pixel resolution, speed, robustness against high illumination background, etc.) and have led to a large number of different patterns to be used in classic structured light methods. Our goal here is to evaluate the potential of three different codification methods to be used in pair with an ATIS sensor. The first two methods use spatio-temporal coding, one using binary code generated from blinking spots with fixed frequency and the second, using caracterizable shapes as code letters. The last method uses moving stripes and compute an unwrapped phase for every pixel in the image (Fig.~\ref{fig:pattern_examples}).

\begin{figure}[!ht]
\centering
\includegraphics[width=1\columnwidth]{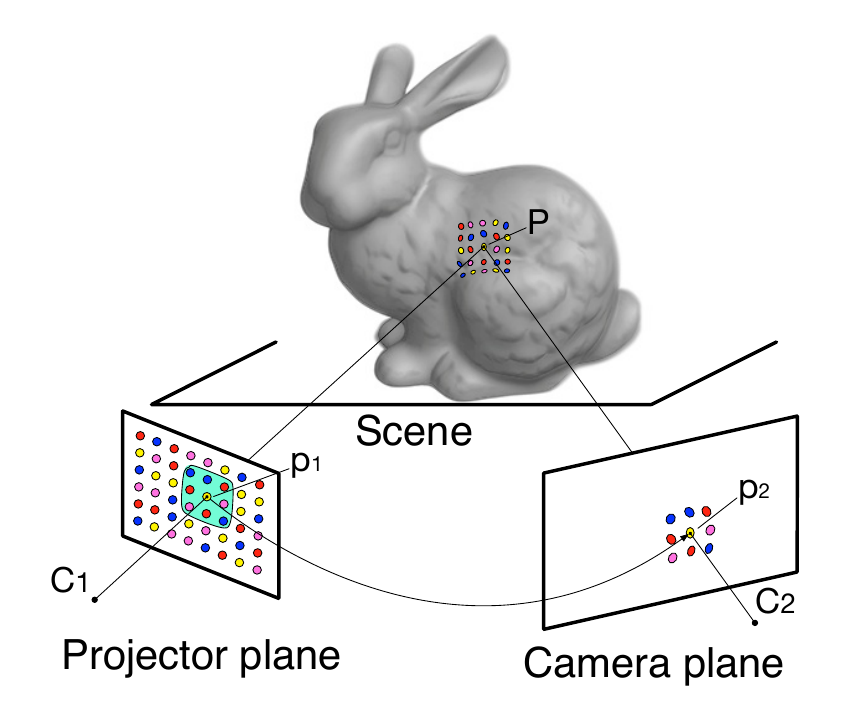}
\caption{Structured light principle. A coded light pattern is projected onto the scene that is observed by a camera. Decoding the pattern allows the matching of paired points in the two views ($p_1, p_2$) and by triangulation of the rays coming from optical centers $C_1$ and $C_2$, the position of the real point $P$ is found. Colors in the picture are for clarity only and in our case refers to different binary signal's dutycycle for the first method or different motion orientation for the second.}
\label{fig:lapin}
\end{figure}

\section{Period and Dutycycle}

As the ATIS sensor is able to perceive really fast changes in illumination, it is possible to exploit the temporal redudancy of a blinking pattern. This approach can be viewed as an extension of standard binary time-multiplexing methods with the specificity that the different code-words are given by the frequency (and dutycycle) of the signal projected onto a pixel instead of being computed through a sequence of intensity values. Fig.~\ref{fig:dutycycle} shows the pattern spatial organisation, each color refers to a specific dutycycle. It uses a Debruijn sequence that is repeated for each line. Ideally we should have projected full lines instead of small spots but practical limitations limited our choices. A periodic projection generates, at the pixel level, an event stream composed of successive ON and OFF event bursts following the signal's periodic behavior. The length and the number of events constituting each burst is defined by the contrast and illumination condition as well as the set of biases used to operate the sensor.

In a perfect case, each edge of the signal would trigger only one event of the corresponding polarity for each pixel. In this situation, measuring the inter-event time between successive ON and OFF events would be enough to get a correct estimation of the signal's frequency and dutycycle. However, because of small imprecisions in the attribution of timestamps by the sensor's arbiter and possible variations in the global illumination, the number of events triggered at each edge of the signal is not consistent. To ensure at least one event per period, we have two choices: increase the sensor's sensibility to generate bursts at each edge of the signal so that faulty events are less prevalent, or consider a spatial neighborhood instead of a single pixel. The first solution has a major drawback as increasing the number of event at high speed may generate congestion in the event stream and results in less accurate timestamps attribution. The second solution, even if lowering the spatial resolution, is more appropriate. 

%% \begin{figure}[!ht]
%% \centering
%% \includegraphics[width=1\columnwidth]{on-off_thing}
%% \caption{Period extraction principle. Each pixel may receive different illumination or have small differences in latency, causing the analysis from a single pixel to be unaccurate, but spatial neighboring allows for a more robust estimation. Applying a burst filter on the neighborhood event stream, we generate two filtered streams, one for each polarity, with precise timings from which we can easily extract both the period and the dutycycle of the signal.}
%% \label{fig:dutycycle}
%% \end{figure}

As a result, instead of considering event streams from each pixel, we create a set of two streams (one for each polarity) composed of events from a local pixel neighborhood of size $\mathcal{N}$. This operation increase the probability of having at least one event per edge but also increase the size of bursts. Eventually, using this spatial neighborhood also increase the precision of the timestamp of the first event in each burst as it is always the pixel with the smallest latency that shoot first, reducing the impact of electronic noise in the sensor's chip.

\subsubsection{Burst filter}

\begin{figure}[!ht]
\centering
\includegraphics[width=1\columnwidth]{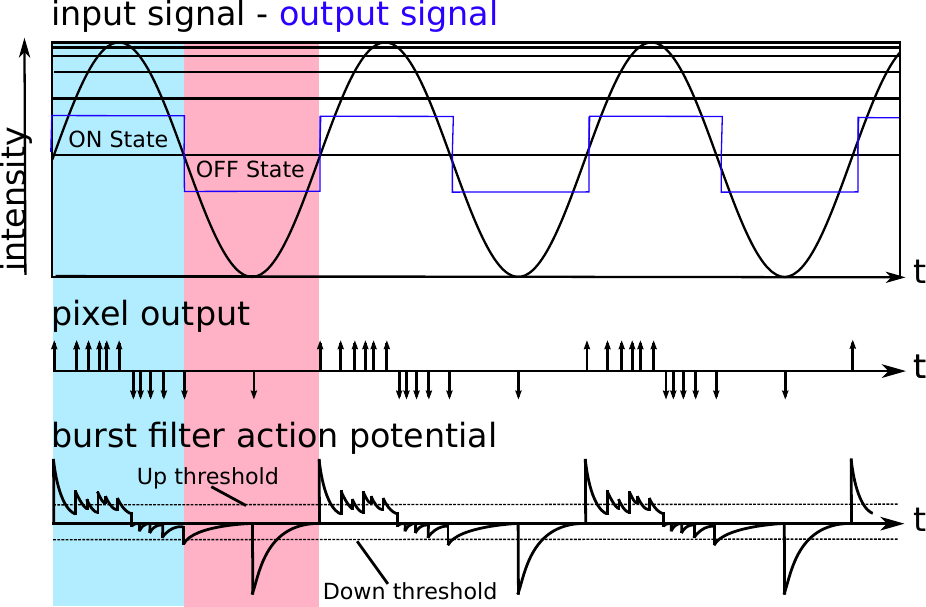}
\caption{\label{fig:burst_filter} Example of the burst filter on a sinusoidal input. The input signal is transformed into an event stream by the sensor and at each new event the inter-spike delay is added to the filter's action potential. Between events, the AP follows an exponential decay. At any given polarity state, when the opposite polarity threshold is reached, the output toggles, effectively reflecting the changes of the input signal. }
\end{figure}

As we are working on binary pattern, every information related to variations of higher frequency than the signal's one has to be removed. The first step in this process is to reduce the sensor's output, from a series of event clusters following the light variations, to a single event at each edge of the input signal. To recover a robust estimation of the signal period (and dutycycle), it is necesssary to clean the bursts and extract, a single, precise event with the correct polarity. A filter is assignated to each pixel's output that integrates inter-event delays as shown on Fig.~\ref{fig:burst_filter}. The filter behaves like two integrate and fire neuron models inhibiting each other but modelized as a single filter, instead of having each neuron triggering when their threshold is reached, it uses a hysteresis function to generate outputs from a single action potential. The filter's action potential is sent through a hysteresis comparator with two different thresholds that enables the control of the sensor dynamics for each polarity. Beginning in an uncertain state at time $t_n$, when a first threshold $thresh(up,down)$ is reached the output state is changed accordingly and it remains untill the opposite polarity threshold is crossed. The variations of the filter's action potential $AP$ then follows the rule: for an event $e_i$  arriving at time $t_{n+1}$ with polarity $p(e_i)$, first the exponential decay is computed as
\begin{equation}
\widehat{AP}(t_n) = AP(t_n)\exp{\frac{(t_{n+1}-t_n)}{\tau}}
\end{equation} 
with $\tau$ a time constant set according to the stimulus frequency, then the update is done by
\begin{equation}
AP(t_{n+1}) = \widehat{AP}(t_n)\cdot(1 + (t_{n+1}-t_n).p(e_i)),
\end{equation} 
where $p(e_i) = \{-1,1\}$.
Filter's output state $S_{out}(t)$ is modified when the following condition is met:
\begin{equation}
S_{out}(t_{n+1}) = |AP_{n+1}| > thresh(\bar{S}_{out}(t_n)),
\end{equation}
where $thresh(\bar{S}_{out}(t_n))$ is the absolute value of the opposite polarity threshold for a given output state. Each time this condition is reached, a new event $e(x,y,t,p)$ is triggered and sent to the rest of the filtering chain with $(x,y)$ as the position of the pixel, timestamp $t_{n+1}$ at which the output state changes and the polarity $p$ reflecting filter's new state. 
The output stream of this filter is a spike train $S_{e(x,y,t,p)}$ with accurate timestamps from which we can easily estimate the signal's frequency $\widehat{f}(x,y)$ after each spike of same polarity by 
\begin{equation}
\widehat{f}(x,y) = \frac{1}{\widehat{T}} = \frac{1}{t_k^p - t_{k-1}^p}.
\end{equation}
The signal's dutycycle is estimated from the computation of its two half periods by comparing the successive alternations of positive and negative filtered events as
\begin{equation}
\widehat{T}_k^{ON} = (1-\lambda) \widehat{T}_{k-1}^{ON} + \lambda (t_k^{OFF} - t_ {k-1}^{ON}),
\end{equation} 
\begin{equation}
\widehat{T}_k^{OFF} = (1-\lambda) \widehat{T}_{k-1}^{OFF} + \lambda (t_k^{ON} - t_{k-1}^{OFF}),
\end{equation}

with $\lambda$ a smoothing parameter.

The signal's dutycycle is given by
\begin{equation}
\widehat{\alpha} = \frac{\widehat{T}^{ON}}{\widehat{T}^{ON}+\widehat{T}^{OFF}}.
\end{equation}

\begin{figure}[!ht]
\centering
\includegraphics[width=1\columnwidth]{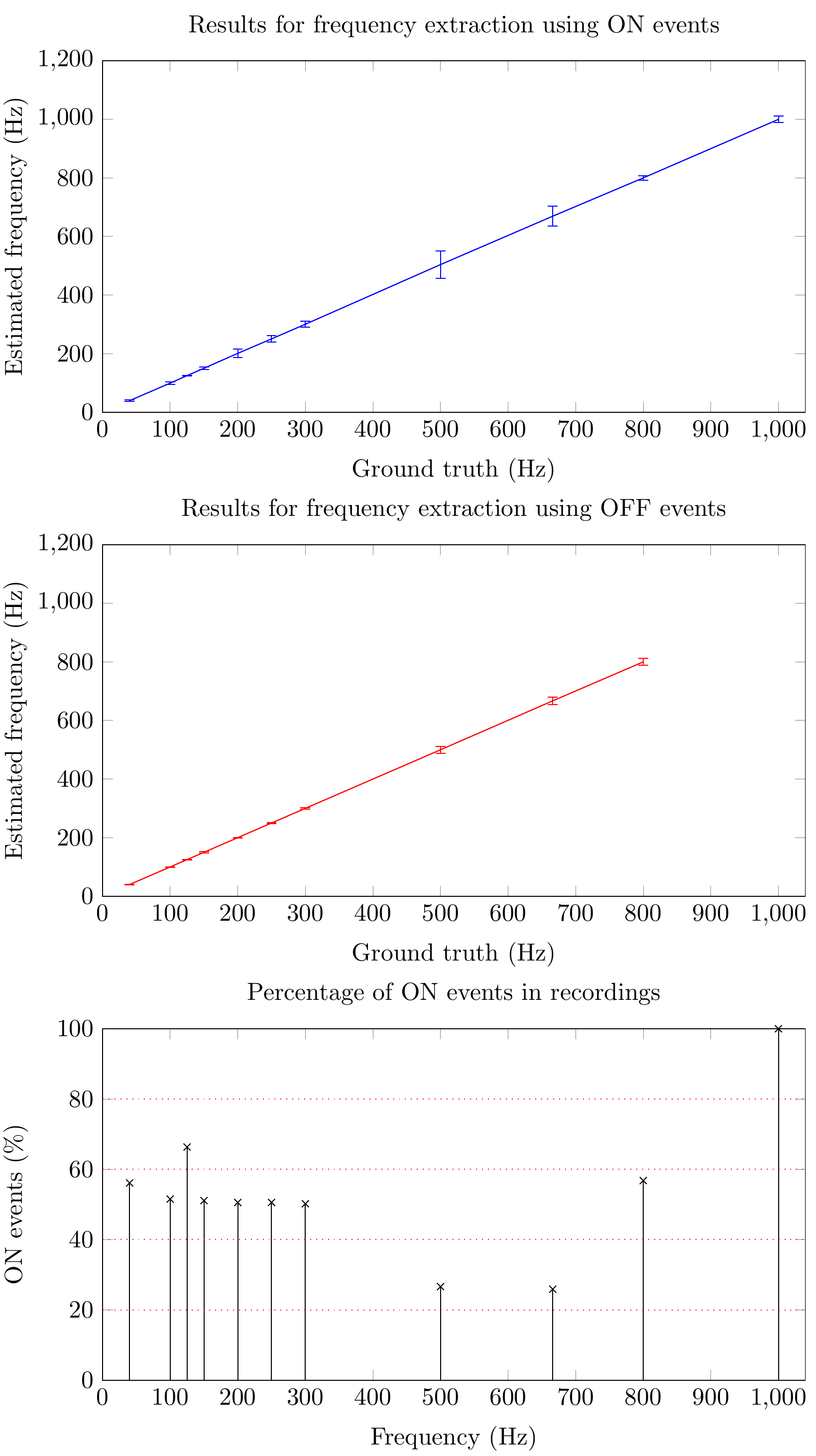}
\caption{Results for frequencies ranging from $40\hertz$ to $1\kilo \hertz$ over 100 periods. Error bars shows the standard deviation observed in each set. Dutycycle was fixed at $50\%$. The third graph shows the precentage of ON events in each set and can explain variations of the variance observed at $500\hertz$ and $666\hertz$.}
\label{fig:TIKZ_f_results}
\end{figure}

Results of the method are shown in Fig.~\ref{fig:TIKZ_f_results} for a wide range of frequencies going from $40\hertz$ and up to $1\kilo \hertz$. Bias settings of the sensor were fixed during the experiment and chosen to cover the complete range of frequencies without changing the behavior of event bursts to reduce the influence of background noise, however the threshold for the OFF channel was a bit too low at $1\kilo \hertz$ and no output could be recorded. This doesn't mean that the algorithm can't extract frequencies over $1\kilo \hertz$, but such high speed requires a particular bias setting. The scope of this paper is to present a pattern design for structured light reconstruction and as such, bias settings should be set in order to optimize the behavior in a narrower range of frequencies around the desired projection rate. 

Error bars shows the standard deviation over the $100$ periods measured at each frequency. Looking at the ON events results for $500\hertz$ and $666\hertz$, we can see that the consistency of the frequency estimation is heavily dependent on the good quality of the data recorded: at those two frequencies, the number of ON events recorded fell to only one third of those of negative polarity, which translates into missed edges or inaccurate timestamps on the filter's ouput. The reason of this drop in events number at particular frequencies is yet unknown but could lay in ambient light condition or background noise. The OFF channel offers more consistency and we can observe a slight increase in variance for high frequencies that express the progressive influence of small imprecision in timestamp attribution as the period decrease closer to the sensor's temporal resolution. 

\begin{figure}[!ht]
\centering
\includegraphics[width=1\columnwidth]{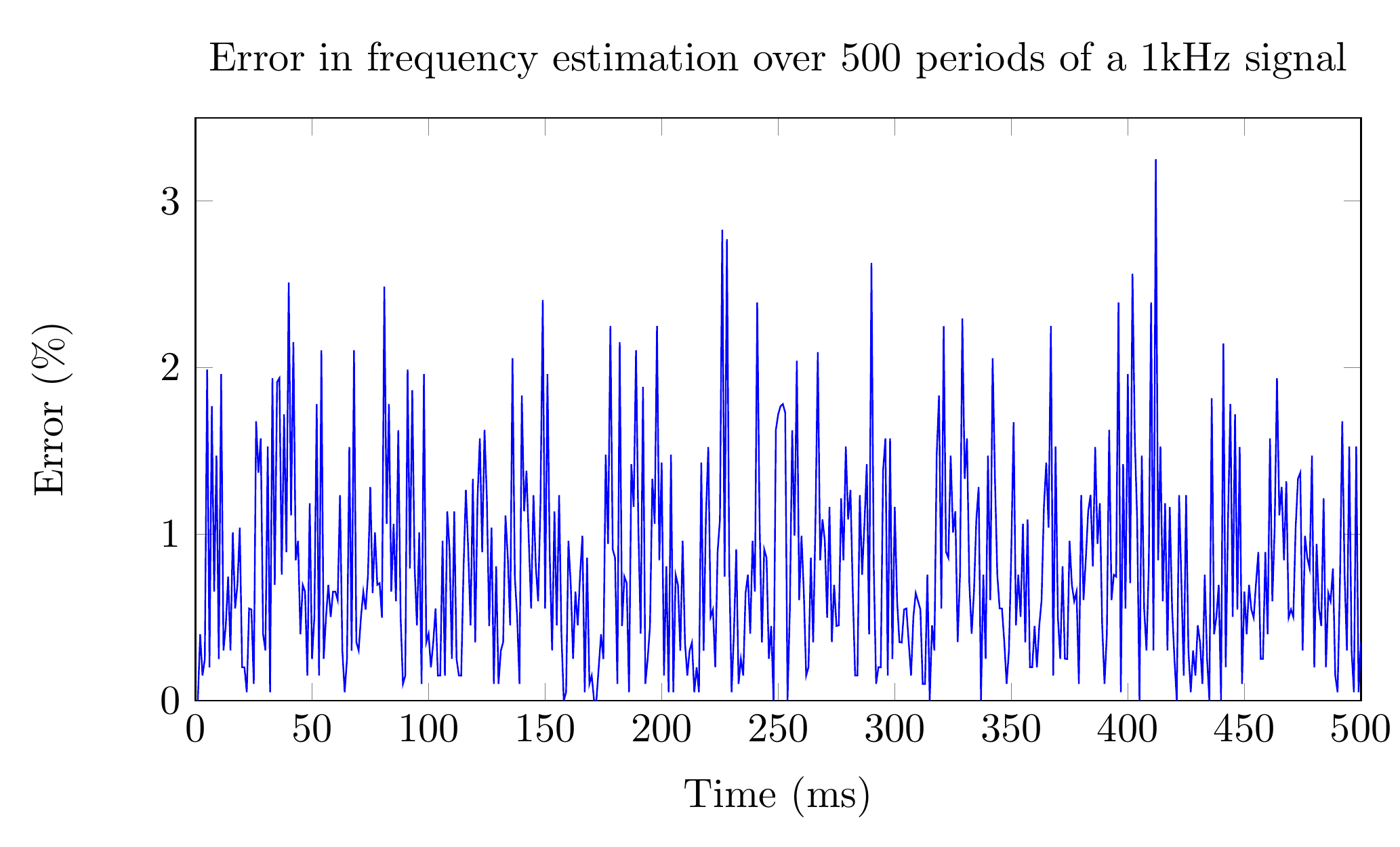}
\caption{Error percentage in frequency estimation, for $500$ periods of a signal at $1\kilo \hertz$ in a $3 \times 3$ pixel neighborhood. The algorithm gives an estimation maintained under less than $3\%$ error.}
\label{fig:error_1kHz}
\end{figure}

In Fig.~\ref{fig:error_1kHz} we show the error in percentage of the projected frequency, obtained for a signal blinking at $1\kilo\hertz$, over $500$ periods. We obtain a mean extracted frequency $\widehat{f} = 999.92\hertz$ for a $3 \times 3$ neighborhood and the error stays under $3\%$, whitch corresponds to a standard deviation of $\sigma = 9.19\hertz$.

\subsection{Random shifting}
Given a set of $n$ signals following a square wave of fixed frequency, with no other treatment done, every signals switch at the same time and will generate at least $2n$ events (ON and OFF alternance) for each period $T$ if used in projection on the ATIS. What we desire is an asynchronous stimulus. Here, not only it is totally synchronous, but it is also eventually a large number of changes for a single time step. We need to lower this number as much as possible to stay under practical limitations and generate smooth activity on the sensor. To ensure that, we introduce a random phase shift for every pixels in the set with uniform distribution so that the probability $p$ of having a change event $X$ in a time window of size $\Delta t$ is

\begin{equation}
\begin{split}
p(X)&=p(t\leq X\leq t+\Delta t) \\
&=\int_t^{t+\Delta t}\frac{2}{T}d\tau \\
&=\frac{2\Delta t}{T}
\end{split}
\end{equation}

The number of changes generated by the whole set of phase shifted pixels follows a binomial distribution $\mathcal{B}(n,p)$ with average value $\mu=np$ and variance $\sigma^2=np(1-p)$ as shown in Fig.\ref{phasehist}.

In a coded pattern constructed from an alphabet of size $k$, we have $k$ sets of pixel with same frequency so the distribution of the total number of changes would be the sum of each binomial functions $\mathcal{B}_k(n_k,p_k)$. An approximation of this sum can be found by a normal distribution $\mathcal{N}(\mu,\sigma)$ as long as $\sigma^2$ is not too small with 

\begin{equation}
\mu=\sum{n_kp_k}
\qquad
\sigma^2=\sum{n_kp_k(1-p_k)}
\end{equation}

\section{Orientation tracking}
This second method aims to take advantage of the powerful tracking capabilities of the ATIS sensor as it has been demonstrated in \ref{Lagorce2014}. Instead of using only time to code the symbols of the pattern, a spatio-temporal code is designed. Spots are moved along different oriented axis in a pattern organized in the same way than the previous method. Tracking is done using gaussian kernels designed in the previously cited paper and, for each determined position, orientation of the moving axis is computed from the covariance matrix by determining the highest correlation axis. Compared to the first method's codification, tracking enables to gather information much faster, as each successive projection step improve the current tracker covariance instead of having to wait a complete period. The acquisition can be made each time the trackers are updated and this delay can be set lower than a complete movement revolution. However, the spatial resolution is lowered by the size of the moving object and symbol density cannot be too high or the tracking would fail to separate symbols from each other.

\subsection{Perfect Maps}
Perfect Maps, are random arrays of dimensions $r \times v$ in which a sub-matrix of dimensions $n \times m$ appears only once in the whole pattern. Theoretically, PM are constructed having dimensions $rv = 2^{mn}$ but usually the zero-matrix is not considered giving $rv = 2^{mn}-1$ unique sub matrices of $m\times n$ dimensions.
In \cite{}, Morano defines a brute force algorithm (non De Bruijn based) to generate these maps by adding successive pseudo-random elements to the pattern, row by row starting from the top left corner. Each new element corresponds to a new sub matrix with an associated code with alphabet of size $k$, and its Hamming distance between all codes already included in the pattern is calculated. If this distance is superior or equal to the minimum Hamming distance specified, the test is passed and the code is added, otherwise the other $k-1$ symbols are tested. When the algorithm gets stuck, the whole pattern is canceled and the algorithm starts over. Since the constructed pattern can contains only a subset of all possible combinations, such PM is called Perfect SubMaps (PSM). Other works used this brute force approach like Claes \cite{} using a colour codification or Albitar \cite{} with geometrical features.
More recently, Maurice and al. \cite{} proposed a hybrid algorithm to build large patterns of more than $200\times 200$ features in a very short time with high Hamming distance, using a splitting strategy to perform the Hamming test in the codeword space instead of the pattern array.
We choose to use in this paper an algorithm inspired by this last method.

To generate a matrix M, with a number of symbols $k$ and a codeword length n, we first create an array with $k^n$ boolean flags for all possible codewords, initialised to zero. Starting on the top left corner of M, we randomly generate a first $3 \times 3$ patch and perform a Hamming test on every codewords with a distance $H < H_{min}$. Each flag array entry is accessed by a single codeword such that the $n$ symbols are the coefficients in $k$ basis.
If all the corresponding flags are set to \textit{false}, the patch is added to M and a next patch is generated until the map is complete (Fig.\ref{fig:PMgen}).
If the algorithm gets stuck in a situation were the added symbol generates only forbidden codes, we shuffle recently added codes until all neighboring codes affected by this modification pass the Hamming test.

%\begin{figure}[!ht]
%\centering
%\includegraphics[width=0.8\columnwidth]{psmap}
%\caption{Algorithm used to generate a Perfect SubMap}
%\label{fig:PMgen}
%\end{figure}

\section{Phase shifting}
Phase shifting is a well known method since 1966 (Carré, 1966) that has been widely used in interferometry for its better accuracy and speed as opposed to static images analysis methods. PS algorithms records a series of images that encode the wavefront phase in the variations of the intensity pattern. Various algorithms has been developped over the years using different number of images and phase unwrapping methods. Similarly, fringe projection for 3D shape reconstruction has been exhaustively studied because of its simple setup and processing associated with relatively high speed and resolution and among those, phase-shifting methods have proven very effective. In standard imaging, intensity values of successive shifted patterns are compared point-to-point to recover a unique phase value for each parrallel line along the coding axis. The existence of a depth discontinuity in the scene will cause a deformation of the projected pattern that will be recorded as a phase deviation as shown on Fig.~\ref{PSexample}. The unique phase value recovered is matched with the projected pattern and the shape of the scene objects can be computed.

\begin{figure}
\centering
	\includegraphics[width=\columnwidth]{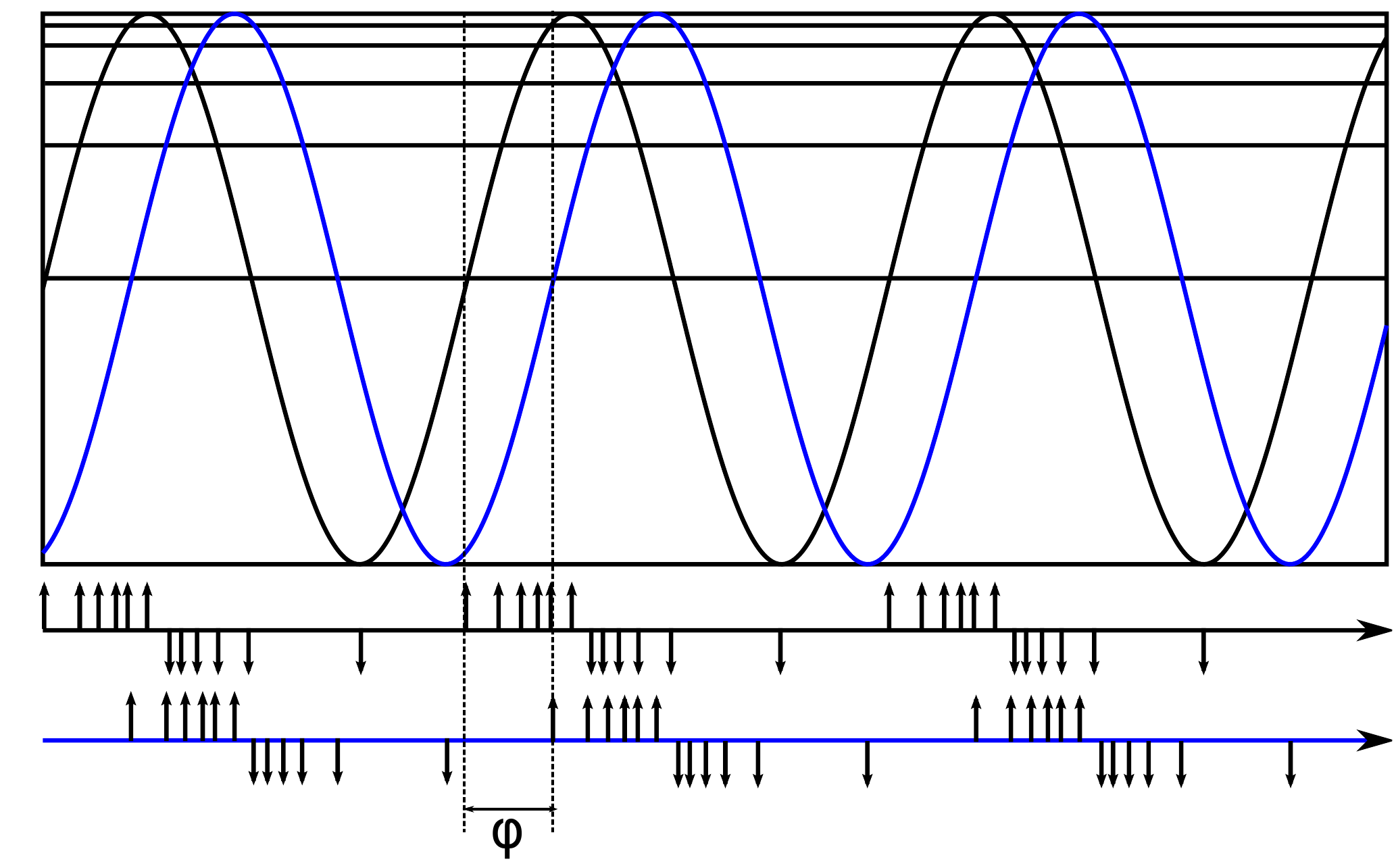}
\caption{\label{phaseshift} When a periodic signal is continusously shifted in front of the sensor, the event stream of two different pixels gives precise temporal information about the phase of the signal.}
\end{figure}

\subsection{Event-base phase-shifting
}
Commonly, in order to recover phase deviation, several phase-shifted patterns are projected and recorded. Using a set number of shifted images, a system of equations is defined and solving it allows to retrieve the phase value. With AES sensors however, no intensity value can be measured directly and no complete frames can be acquired which makes common methods unusable. Instead, a continuous shifting of the pattern is used to produce on each line along the coding axis, a series of events with precise timestamps following the variations of the pattern fringes. Every period of the signal projected generates on each pixels a succession of alternatively positive and negative spike bursts $\mathcal{S}(p)$ as shown in Fig.~\ref{phaseshift}. Taking a given pixel $p_i$ as reference on a line, for every pixels $p_{j,\forall j\neq i}$, we can measure the time-shift between bursts $\mathcal{S}(p_j)$ and $\mathcal{S}(p_i)$. To recover the wrapped phase $\phi$ , a modulo operator is applied on this value to constrain it between 0 and the signal's period, then it is projected on the interval $[0,2\pi)$ by :
\begin{equation}
\phi = \frac{2\pi }{P_{signal}}mod( \Delta t(\mathcal{S}(p_j),\mathcal{S}(p_i)), P_{signal} )
\end{equation}
An unwrapping algorithm is used to compute the absolute phase value $\Phi$ for every pixels. On each line along the coding axis, the following reccursive formula is applied :
\begin{equation}
\Phi _{x} = \Phi _{x-1} + \lambda * mod(\phi _{x-1} - \Phi _{x-1}, 2\pi)
\end{equation}
Once the unwrapped phase is obtained, a mapping between the projected pattern and the phase image can be found and the 3D measurement is done.

%%%%%%%%%%%%%%%%%%%%%%%%%%%%%%%%%%%%%%%%%%%%%%%%%%%%%%%%%%%%%%
\section{Results}
%%%%%%%%%%%%%%%%%%%%%%%%%%%%%%%%%%%%%%%%%%%%%%%%%%%%%%%%%%%%%%
\label{section:results}

\subsection{Experiment}

\begin{figure}[!ht]
\centering
\includegraphics[width=0.8\columnwidth]{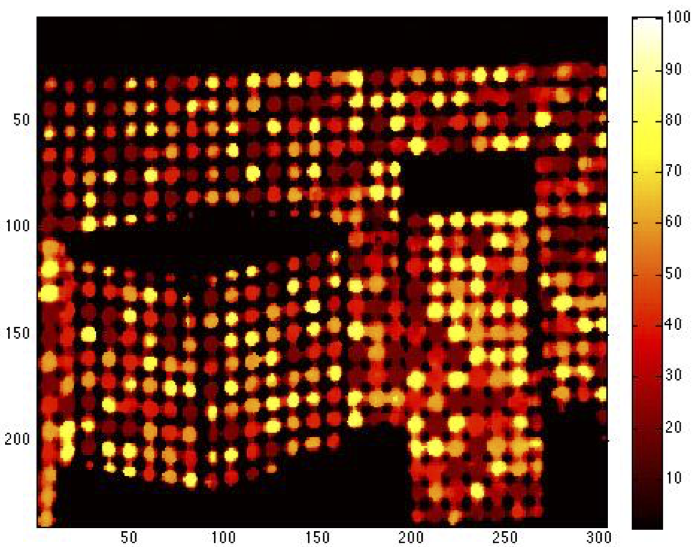}
\caption{Image of the projected pattern as extracted by the algorithm. Colors correspond to the estimated dutycycle.}
\label{fig:DCim}
\end{figure}

\begin{figure}[!ht]
\centering
\includegraphics[width=0.8\columnwidth]{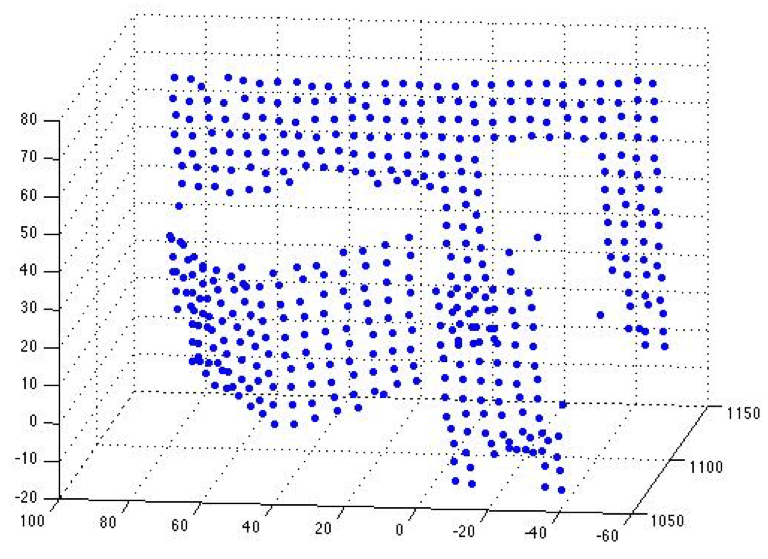}
\caption{3D point cloud.}
\label{fig:3Dim}
\end{figure}

The method is applied on a scene containing two objects placed around one meter away from the system. We used a SPSM of $20 \times 30$ symbols and the pattern codification is done using the dutycycle of the signal. The neighborhood in the burst filter is set to 1 ($3 \times 3$ pixels) and signal's frequency is set to $20 \hertz$ for performance issues. The algorithm runs on MatLab and even if it could be heavily parallelized, it wasn't the case on this experiment and real time wasn't achieve. An image of accumulated outputs from filters is given in Fig.~\ref{fig:DCim}, colors in the image reflect the estimation of the dutycycle by each pixel associated filter. Point correspondance is performed by extracting the $3 \times 3$ dutycycle based codewords and as pattern's dots are bigger than a single pixel, we perform a spatial averaging to obtain a sub-pixel position of the dot in the camera frame. After triangulation, a 3D point cloud is obtained (Fig.~\ref{fig:3Dim})

\begin{figure}[!ht]
\centering
\includegraphics[width=0.8\columnwidth]{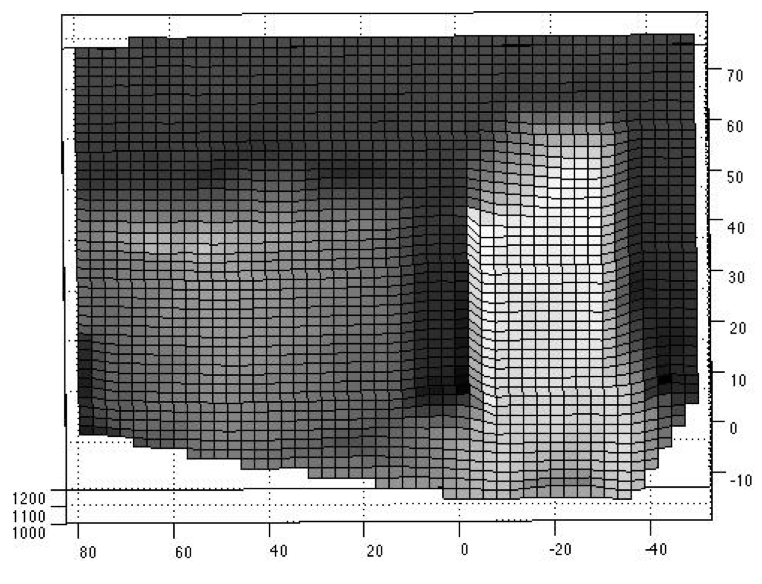}
\caption{Reconstructed depth}
\label{fig:fitim}
\end{figure}

%%%%%%%%%%%%%%%%%%%%%%%%%%%%%%%%%%%%%%%%%%%%%%%%%%%%%%%%%%%%%%
\section{Conclusion and Discussion}
This paper introduced a methodology that makes use of the hguh temporal resolution of the event based sensor to rethink the problem of structured light depth reconstruction. We used a unique spatial  distribution of light patterns composed of elements each flickering at a unique frequency. We also used the idea of random shift allowing each neighboorhood to be unique. Each methodology relying on the combined use of an event based camera and a light projector coding each spatial position in the frequency domain can make use of the developed approach. There are several ways to decode frequencies from the output of the event based camera. The method used here could be bettered, using more precise event based camera allowing fequency to be extracted from the timing of the oscillation of each pattern from inter spike information. If that condition is fulfilled even simpler patterns could be used without the need to rely on unique neighborhoods. 

%%%%%%%%%%%%%%%%%%%%%%%%%%%%%%%%%%%%%%%%%%%%%%%%%%%%%%%%%%%%%%

%\section*{Acknowledgment}


\ifCLASSOPTIONcaptionsoff
  \newpage
\fi

\bibliographystyle{IEEEtran}
\bibliography{Biblio.bib}
%\bibliography{IEEEabrv,../bib/Biblio}
%\enlargethispage{-12cm}

\end{document}